\documentclass[runningheads]{llncs}
\usepackage[T1]{fontenc}
\usepackage{graphicx}
\usepackage[linesnumbered,ruled,vlined]{algorithm2e}
\usepackage{multirow}
\usepackage{float}
\usepackage{tabularx}
\usepackage{subcaption}
\usepackage{orcidlink}
\usepackage{amsfonts}
\begin{document}
\title{MapFM: Foundation Model-Driven HD Mapping \\ with Multi-Task Contextual Learning}
\titlerunning{MapFM}
%
\author{Leonid Ivanov\inst{1}\orcidlink{0009-0008-1748-1804} 
\and
Vasily Yuryev\inst{1}\orcidlink{0009-0004-2965-9981} 
\and
Dmitry Yudin\inst{1,2}\orcidlink{0000-0002-1407-2633} 
}
\authorrunning{L. Ivanov et al.}
%
\institute{Moscow Institute of Physics and Technology, 141701, Dolgoprudny, Russia
\and
AIRI, 121170, Moscow, Russia
}
\maketitle              



\begin{abstract}
In autonomous driving, high-definition (HD) maps and semantic maps in bird's-eye view (BEV) are essential for accurate localization, planning, and decision-making. This paper introduces an enhanced End-to-End model named MapFM for online vectorized HD map generation. We show significantly boost feature representation quality by incorporating powerful foundation model for encoding camera images. To further enrich the model's understanding of the environment and improve prediction quality, we integrate auxiliary prediction heads for semantic segmentation in the BEV representation. This multi-task learning approach provides richer contextual supervision, leading to a more comprehensive scene representation and ultimately resulting in higher accuracy and improved quality of the predicted vectorized HD maps.  The source code is available at https://github.com/LIvanoff/MapFM.
\end{abstract}

\section{Introduction}  

\label{sec:intro}
High-definition (HD) semantic maps are among the most important components of autonomous driving technology. The use of semantic information about the environment, such as lane boundaries, road dividers and pedestrian crossings, plays a crucial role in vehicle navigation, motion prediction, and planning. Traditional methods of HD map construction, such as Simultaneous Localization and Mapping (SLAM) and manual map annotation, require significant time and human effort. These limitations are driving a shift towards online, learning-based methods that leverage onboard sensor data.

\begin{figure}[t]
  \centering
   \includegraphics[width=1.0\linewidth]{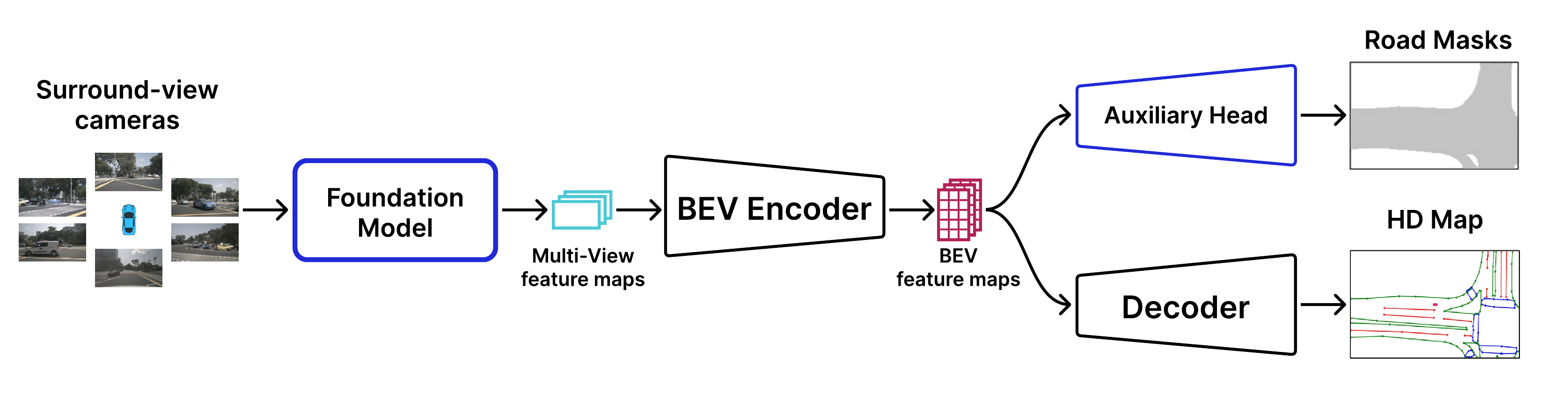}
   \caption{The general pipeline of the proposed MapFM method with foundation model predicts a vector HD map.}
   \label{fig:pipeline}
\end{figure}

A pioneering work in the field of learning-based methods for HD map construction is the HDMapNet~\cite{li2022hdmapnet}, which perform semantic segmentation in BEV space by producing rasterized maps. Several existing methods~\cite{dong2024superfusion,zhou2022cross,wang2023lidar2map,schramm2024bevcar} allow the construct vectorized HD maps through post-processing of rasterized images. However, these approaches often demonstrate inefficiencies, both in the quality of the resulting vectorized maps and in the speed of the post-processing pipeline. Alternatively, recent approaches showcase the benefit of leveraging rasterized maps not as the primary output requiring vectorization, but as an efficient intermediate representation or prior. A notable example is HRMapNet~\cite{zhang2024enhancing}, which effectively utilizes a historical rasterized map as a low-cost source of complementary information. This historical context significantly improves the performance of the main online vectorized map generation module.

To enhance the quality of vectorized map prediction and address the limitations of traditional segmentation methods, recent approaches have shifted towards directly generating sets of points for map construction. These methods~\cite{liao2022maptr,liu2024mgmap,liao2024maptrv2,liu2024leveraging,liu2023vectormapnet,ding2023pivotnet,hu2024admap}, leveraging transformer decoders similar to DETR~\cite{carion2020end}, allow for the immediate generation of structured maps without the need for additional post-processing.

Existing online HD map methods often employ standard backbones like ResNet~\cite{he2015deepresiduallearningimage} and Swin Transformer~\cite{liu2021swintransformerhierarchicalvision}, potentially limiting representation power compared to modern foundation models. We propose leveraging the robust features from large-scale pre-trained models like DINOv2 and RADIOv2.5 to enhance the perception module for map generation. This novel application aims to improve map accuracy by building on a stronger feature foundation.

In this paper, we introduce an efficient framework  designed to significantly improve the accuracy and robustness of online vectorized HD map generation. Our approach makes two key contributions. First, departing from standard task-specific encoders, we harness the representational power of large-scale pre-trained foundation models, namely DINOv2~\cite{oquab2023dinov2} for initial image feature extraction, which, to our knowledge, is a novel application in this domain. Second, to further enhance the model's understanding of the driving environment, we introduce auxiliary heads for BEV semantic segmentation of driveable area on different level.

To summarize, this paper makes the following contributions:
\begin{itemize}
    \item We propose MapFM, an online end-to-end HD map construction framework. It leverages the MapQR query mechanism for efficient map element instance modeling and introduces auxiliary BEV semantic segmentation heads, enabling the model to learn a richer understanding of the scene context through multi-task supervision.
    \item We pioneer the integration and demonstrate the effectiveness of using pre-trained vision foundation model as powerful feature extractors for this task.
\end{itemize}
\section{Related works}
\label{sec:formatting}

\subsection{Online HD map construction}
Online HD map construction seeks to generate maps directly from onboard sensors, mitigating issues of cost and freshness associated with offline methods. Early approache relied on BEV semantic segmentation~\cite{li2022hdmapnet}, but required complex post-processing to obtain vectorized outputs, often losing precision and instance details~\cite{hu2024admap}. This led to the development of end-to-end methods that directly predict vectorized representations.

To address these limitations, subsequent research focused on directly predicting vectorized map elements. VectorMapNet~\cite{liu2023vectormapnet} was a pioneering end-to-end method using a two-stage approach with an auto-regressive decoder to predict points sequentially for each map instance. However, its sequential nature can be slow and potentially prone to error accumulation.

\subsection{Transformer-based HD mapping}
 Inspired by the success of DETR~\cite{carion2020end} in object detection, MapTR~\cite{liao2022maptr} introduced a paradigm shift by modeling map elements as sets of points and using a parallel transformer decoder. This has improved the efficiency and speed of the end-to-end approach. MapTRv2~\cite{liao2024maptrv2} further refined this by introducing decoupled self-attention for efficiency, auxiliary losses for better supervision, and support for directed map elements like lane centerlines.
Other contemporary works have explored different representations or mechanisms. BeMapNet~\cite{qiao2023endtoendvectorizedhdmapconstruction} and PivotNet~\cite{ding2023pivotnet} utilize Bézier curves or pivot points, respectively, for a more compact representation of map elements compared to raw point sequences. Addressing temporal consistency, StreamMapNet~\cite{yuan2024streammapnet} and MapTracker~\cite{chen2024maptracker} introduced streaming architectures that propagate information (BEV features or queries) across time frames. Methods like HRMapNet~\cite{zhang2024enhancing} and GlobalMapNet~\cite{shi2024globalmapnet} leverage historical or global map information (often rasterized or vectorized) as an explicit prior to enhance current predictions. More recently, MapQR~\cite{liu2024leveraging} focused on improving the query design within the DETR-like framework using a scatter-and-gather mechanism for better information probing and instance modeling. 
Our work builds upon these DETR-like approaches, particularly improving the query mechanism and feature representation.

\subsection{Foundation models for image-based mapping}
In recent years, there has been a paradigm shift with the advent of Foundation models. These models are typically large-scale neural networks pre-trained on vast amounts of broad, often unlabeled, data using SSL techniques. Their key characteristic is the ability to serve as a strong "foundation" for a wide variety of downstream tasks, often achieving remarkable performance with minimal task-specific fine-tuning (few-shot or zero-shot learning) or by simply providing powerful, general-purpose feature representations.

Models like CLIP~\cite{radford2021learningtransferablevisualmodels} learned joint image-text embeddings from web-scale data, enabling impressive zero-shot image classification. MAE (Masked Autoencoders)~\cite{singh2024effectivenessmaeprepretrainingbillionscale} and BEiT~\cite{bao2022beitbertpretrainingimage} adapted the masked language modeling idea from BERT~\cite{devlin2019bertpretrainingdeepbidirectional} to vision, learning rich representations by reconstructing masked patches of an image. DINO~\cite{caron2021emergingpropertiesselfsupervisedvision} and DINOv2~\cite{oquab2023dinov2} leverage self-distillation and self-supervised learning to extract transferable visual features for dense prediction tasks without fine-tuning. Pre-training on large, diverse datasets makes them strong general-purpose vision backbones. Previous methods~\cite{Schramm_2024,barın2024robustbirdseyeview} used DINOv2 for semantic segmentation and 3D object detection, in particular the BEVCar~\cite{Schramm_2024} method used a ViT-Adapter for frozen DINOv2.


Previous online HD map construction predominantly used standard backbones like ResNet~\cite{he2015deepresiduallearningimage} and SwinTransformer~\cite{liu2021swintransformerhierarchicalvision}. We propose leveraging features from large-scale foundation models like RADIOv2.5 and DINOv2. To our knowledge, this is the first application of such models as direct feature extractors for online vectorized HD map construction. We aim to improve map accuracy and robustness by building upon this stronger perceptual foundation.
\section{Method}
\label{sec:formatting}

\subsection{Overall architecture}

Our proposed method, MapFM introduces significant enhancements in feature representation and contextual understanding through multi-task learning and the integration of foundation model.

\begin{figure}[t]
  \centering
   \includegraphics[width=1\linewidth]{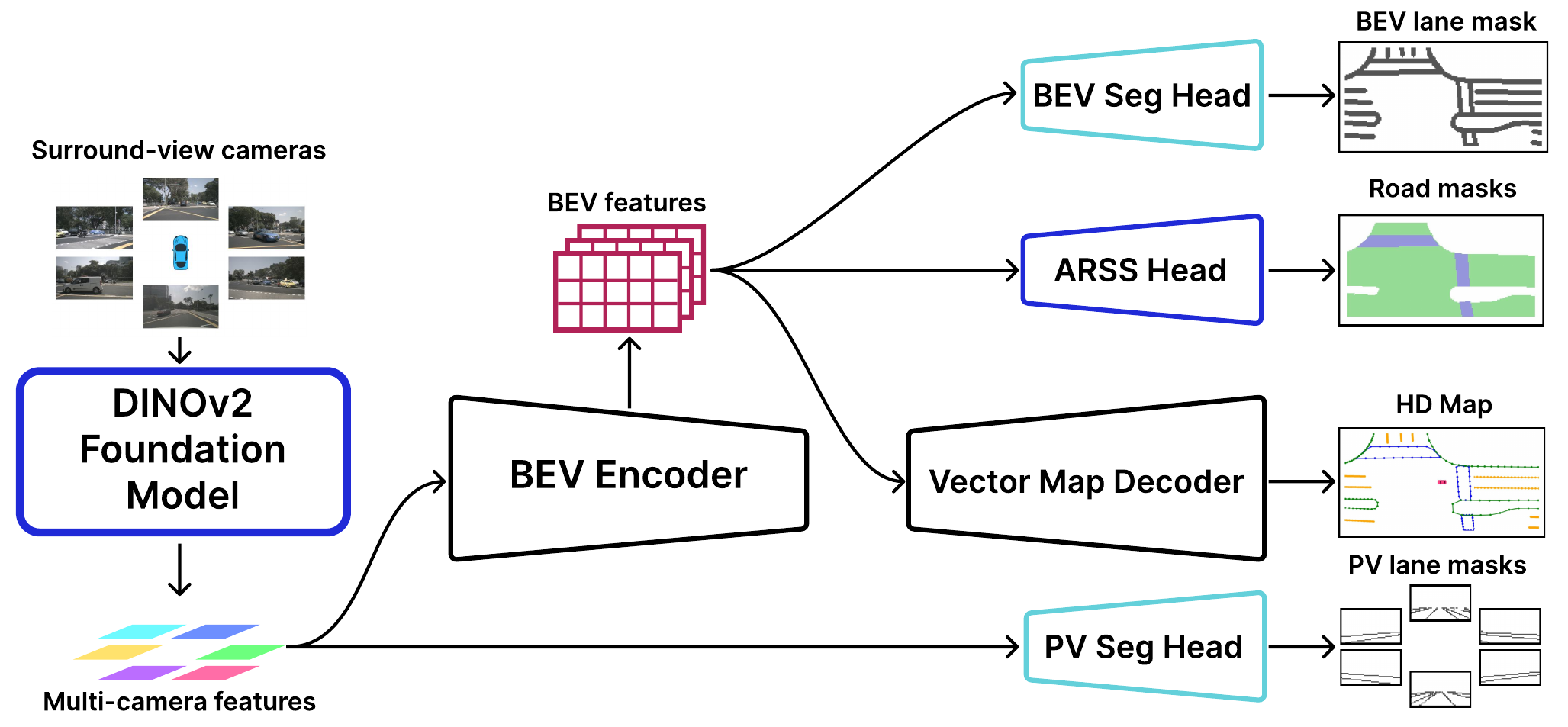}
   \caption{Overview of the proposed MapFM method. Our approach  integrates a Foundation Model and an Auxiliary Road Surface Segmentation Head (ARSS Head) for enhanced environmental perception.}
   \label{fig:mapfm}
\end{figure}

The overall architecture, depicted in Fig. ~\ref{fig:mapfm}, retains the core components: Foundation Backnone (DINOv2) for  feature extraction from multi-view images $\mathcal{I}=\{I_1, \ldots, I_M\} $, a BEV Encoder for transform milti-camera features $\mathcal{F}=\{F_1, \ldots, F_M\} $ to BEV representation $\mathcal{B} \in \mathbb{R}^{C\times H \times W}$, where C, H, W represent the feature channels, height, and width of the BEV feature, a Auxiliary Road Surface Segmentation Head (ARSS Head), a Vector Map Decoder for map element prediction, and an auxiliry heads for BEV and Perspective View segmentation from MapTRv2~\cite{liao2024maptrv2}.

\subsection{Foundation Model} To capture richer and more generalizable features from surrounding-view RGB images as inputs, we replace standard backbone with powerful pre-trained foundation models. For camera inputs, the conventional ResNet~\cite{he2015deepresiduallearningimage} is substituted with a fine-tuned DINOv2~\cite{oquab2023dinov2} model, leveraging its strong visual representations learned from large-scale self-supervised pre-training. These foundation model serve as the perceptual cornerstone of our method.

\subsection{BEV Encoder}
To transform the multi-camera features $\mathcal{F}$ extracted by the DINOv2 backbone into a unified BEV representation, we employ the BEV Encoder module from BEVFormer~\cite{li2022bevformerlearningbirdseyeviewrepresentation}. This encoder, denoted by a function $\Phi_{Enc}$, aggregates spatiotemporal information from the multi-view features using learnable BEV queries and attention mechanisms. The resulting BEV feature map $\mathcal{B}$ is formally defined as:
\begin{equation}
    \mathcal{B} = \Phi_{Enc}(\mathcal{F}, Q_{BEV}, P_{cam}, E_{cam}),
\end{equation}
where $Q_{Enc}$ represents the learnable BEV queries, $P_{cam}$ are the camera projection matrices, and $E_{cam}$ are the camera extrinsic parameters. The output $\mathcal{B}$ serves as the input for both the Auxiliary Road Surface Segmentation Head and the Vector Map Decoder.

\subsection{Auxiliary Road Surface Segmentation Head}
\begin{figure}[H]
  \centering
   \includegraphics[width=1\linewidth]{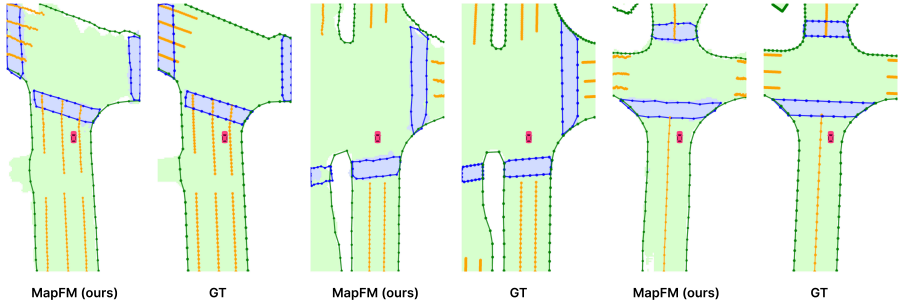}
   \caption{Visualization of HD maps overlaid over road surface segmentation masks. \textbf{Green} -- drivable area mask, \textbf{Blue} -- pedestrian crossing area mask}
   \label{fig:mapfm_combined}
\end{figure}

Building upon the BEV features $\mathcal{B}$ from the BEV Encoder, we introduce an auxiliary segmentation head, denoted by $\Phi_{ARSS}$. This head is designed to perform dense semantic segmentation in the BEV space. Specifically, it predicts two types of BEV masks: a drivable area mask $\mathcal{M}_{drive}$ and a pedestrian crossing area mask $\mathcal{M}_{ped}$:
\begin{equation}
    \mathcal{S}_{BEV} = \Phi_{ARSS}(\mathcal{B}) = \{\mathcal{M}_{drive}, \mathcal{M}_{ped}\},
\end{equation}
where $\mathcal{M}_{drive}, \mathcal{M}_{ped} \in \{0,1\}^{H \times W}$. Training this head with appropriate segmentation losses on these masks provides rich contextual supervision, implicitly aiding the primary task of vectorized map prediction. Examples of such masks are shown in Fig. \ref{fig:mapfm_combined}.

Furthermore, to enrich the learning signals for backbone and BEV encoder, we also incorporate auxiliary heads for perspective view lane segmentation and BEV segmentation, similar to the multi-task learning approach in MapTRv2~\cite{liao2024maptrv2}. These additional tasks ensure that the ours feature extractor and encoder learns a comprehensive set of features relevant to various road elements, ultimately benefiting both dense and vectorized map predictions.

\subsection{Vector Map Decoder}
For the primary task of generating vectorized HD map elements, we retain the core MapQR decoder architecture~\cite{liu2024leveraging}, denoted by $\Phi_{Dec}$. This decoder takes the contextually enriched BEV features $\mathcal{B}$ and a set of $N$ learnable instance queries $Q_{inst}$ to predict the final vectorized map $M$. The map $M$ is represented as a set of $k$ polylines, where $k \le N$:
\begin{equation}
    M = \Phi_{Dec}(\mathcal{B}, Q_{inst}) = \{(P^1, c^1), (P^2, c^2), \dots, (P^k, c^k)\},
\end{equation}
Each polyline $P^i$ is defined as an ordered sequence of $n_i$ 2D points $P^i = \{p_1^i, p_2^i, \dots, p_{n_i}^i\}$, where $p_j^i \in \mathbb{R}^2$. Each $c^i$ is a class label (e.g., road boundary, lane divider, pedestrian crossing) associated with the polyline $P^i$. The decoder utilizes its efficient scatter-and-gather query mechanism and positional embeddings for this prediction.

\subsection{Loss Function}
The overall loss function $\mathcal{L}_{total}$ is a weighted sum of several components. We adopt the point loss $\mathcal{L}_{pts}$, classification loss $\mathcal{L}_{cls}$, direction loss $\mathcal{L}_{dir}$, general BEV segmentation loss $\mathcal{L}_{BEVseg}$, and perspective view segmentation loss $\mathcal{L}_{PVseg}$ from MapTRv2~\cite{liao2024maptrv2} and MapQR~\cite{liu2024leveraging}.

In addition, we introduce a specific surface segmentation loss $\mathcal{L}_{surf}$ for our Auxiliary Road Surface Segmentation Head. This loss is applied to the predicted set of semantic BEV masks $\mathcal{S}_{BEV}$ and their corresponding ground-truth masks $\mathcal{S}^{GT}_{BEV} = \{\mathcal{M}^{1,GT}_{seg}, \dots, \mathcal{M}^{K_{seg},GT}_{seg}\}$. $\mathcal{L}_{surf}$ is formulated as a sum of Dice Losses for each of the $K_{seg}$ semantic classes:
\begin{equation}
    \mathcal{L}_{surf}(\mathcal{S}_{BEV}, \mathcal{S}^{GT}_{BEV}) = \sum_{j=1}^{K_{seg}} \mathcal{L}_{Dice}(\mathcal{M}^j_{seg}, \mathcal{M}^{j,GT}_{seg}),
\end{equation}
where $\mathcal{L}_{Dice}$ is the standard Dice Loss, chosen for its effectiveness in handling potentially imbalanced segmentation tasks. 

The total loss is then:
\begin{equation}
\mathcal{L}=\beta_{1}\mathcal{L}_{pts}+\beta_{2}\mathcal{L}_{cls}+\beta_{3}\mathcal{L}_{dir}+\beta_{4}\mathcal{L}_{BEVseg}+\beta_{5}\mathcal{L}_{PVseg}+\beta_{6}\mathcal{L}_{surf}.
\end{equation}



% ...
\section{Experiments}

We compare the proposed method MapFM with SOTA methods, including MapTR series~\cite{liao2022maptr,liao2024maptrv2}, BeMapNet~\cite{qiao2023endtoendvectorizedhdmapconstruction}, PivotNet~\cite{ding2023pivotnet}, MGMapNet~\cite{yang2024mgmapnetmultigranularityrepresentationlearning} and our baseline MapQR~\cite{liu2024leveraging}. 
\subsection{Settings}
We evaluate our method on public datasets nuScenes~\cite{caesar2020nuscenesmultimodaldatasetautonomous}. These dataset provide surrounding images captured by self-driving cars in hundreds of city scenes, partitioned into 700 scenes comprising 28,130 samples for training purposes, and 150 scenes containing 6019 samples designated for validation. Each sample contains 6 RGB images. NuScenes provides 2D vectorized map elements as ground truth.

\textbf{Evaluation Metric.} Following previous works~\cite{li2022hdmapnet,liao2022maptr}, we evaluate the performance on three static map categories: lane divider, pedestrian crossing, and road boundary. To determine if a predicted map element matches its corresponding ground truth, we utilize the Chamfer Distance (CD). The primary evaluation metric is the mean Average Precision (mAP), calculated based on CD matches at three distinct distance thresholds: 0.5m, 1.0m, and 1.5m, as established in MapTR~\cite{liao2022maptr}.




\textbf{Training and Inference Details.} We follow most settings as in the MapTR series~\cite{liao2022maptr,liao2024maptrv2}, and the modifications will be emphasized. The BEV feature is set to be 200 x 100 to perceive [-30m, 30m] for rear to front, and [-15m, 15m] for left to right. We use 100 instance queries to detect map element instances, and each instance is modelled by 20 sequential points  (We set N = 100 and n = 20.)
Our model is trained on 2 NVIDIA V100 GPUs with the total batch size of 32. The
learning rate is set to $4\times 10^{-4}$. The weights of each loss are set to $\beta_{1}=5$, $\beta_{2}=2$, $\beta_{3}=0.005$, $\beta_{4}=1$, $\beta_{5}=1$ and $\beta_{6}=2$.

\subsection{Comparisons with State-of-the-art Methods}

\begin{table}[t]
\centering
\scriptsize
\caption{Comparison with SOTA methods on nuScenes with thresholds {0.5m, 1.0m, 1.5m}}
\label{tab:sota_compare}
\begin{tabularx}{\linewidth}{|X|X|c|c|c|c|c|} 
\hline
\textbf{Method} & \textbf{Backbone} & \textbf{Epochs} & $\mathrm{AP}_{div}$ & AP$_{ped}$ & AP$_{bound}$ & \textbf{mAP} \\ \hline 
MapTR~\cite{liao2022maptr} & ResNet50 & 24 & 51.5 & 46.3 & 53.1 & 50.3 \\ \hline
MapTR~\cite{liao2022maptr} & SwinT & 24 & 45.2 & 52.7 & 52.3 & 50.1 \\ \hline
BeMapNet~\cite{qiao2023endtoendvectorizedhdmapconstruction} & ResNet50 & 30 & 62.3 & 57.7 & 59.4 & 59.8 \\ \hline
BeMapNet~\cite{qiao2023endtoendvectorizedhdmapconstruction} & SwinT & 30 & 64.4 & 61.2 & 61.7 & 62.4 \\ \hline
PivotNet~\cite{ding2023pivotnet} & ResNet50 & 24 & 56.5 & 56.2 & 60.1 & 57.6 \\ \hline
PivotNet~\cite{ding2023pivotnet} & SwinT & 24 & 60.6 & 59.2 & 62.2 & 60.6 \\ \hline
MapTRv2~\cite{liao2024maptrv2} & ResNet50 & 24 & 62.4 & 59.8 & 62.4 & 61.5 \\ \hline
MGMapNet~\cite{yang2024mgmapnetmultigranularityrepresentationlearning} & ResNet50 & 24 & 66.1 & 64.7 & 69.4 & 66.8 \\ \hline
MapQR (baseline)~\cite{liu2024leveraging} & ResNet50 & 24 & 68.0 & 63.4 & 67.3 & 66.3 \\ \hline
MapQR (baseline)~\cite{liu2024leveraging} & SwinT & 24 & 68.1 & 63.1 & 67.1 & 66.1 \\ \hline
\textbf{(ours) MapFM} & DINOv2-small & 24 & \textbf{68.8} & \textbf{65.7} & \textbf{68.9} & \textbf{67.8} \\ \hline
\textbf{(ours) MapFM} & DINOv2-base & 16 & \textbf{68.8} & \textbf{67.8} & \textbf{71.3} & \textbf{69.0} \\ \hline
\end{tabularx}
\vspace{0.2cm}
\end{table}

\begin{figure}[t]
  \centering
  \begin{subfigure}{0.495\linewidth}
    \centering
    \includegraphics[width=\linewidth]{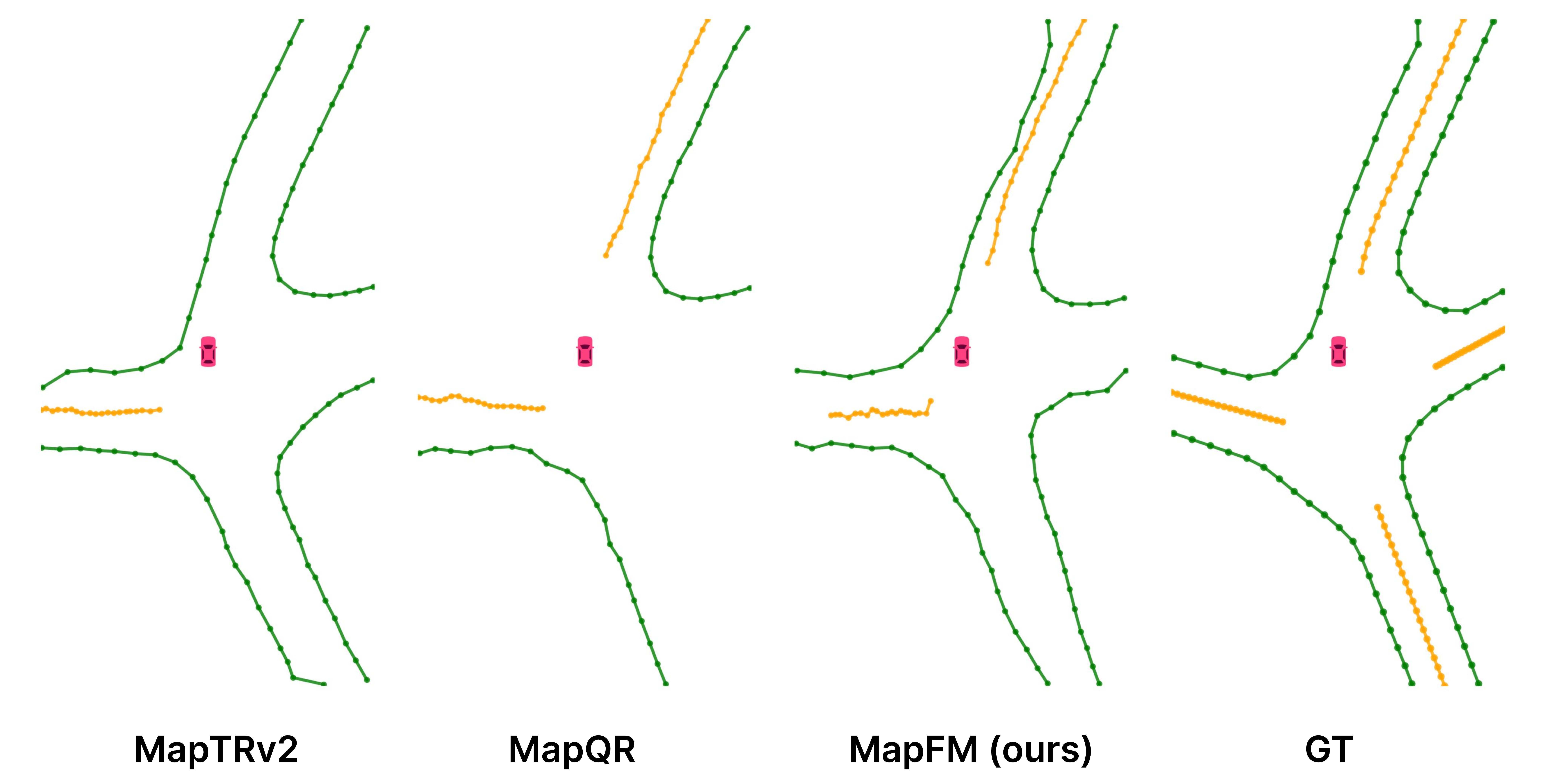}
  \end{subfigure}
  \begin{subfigure}{0.495\linewidth}
    \centering
    \includegraphics[width=\linewidth]{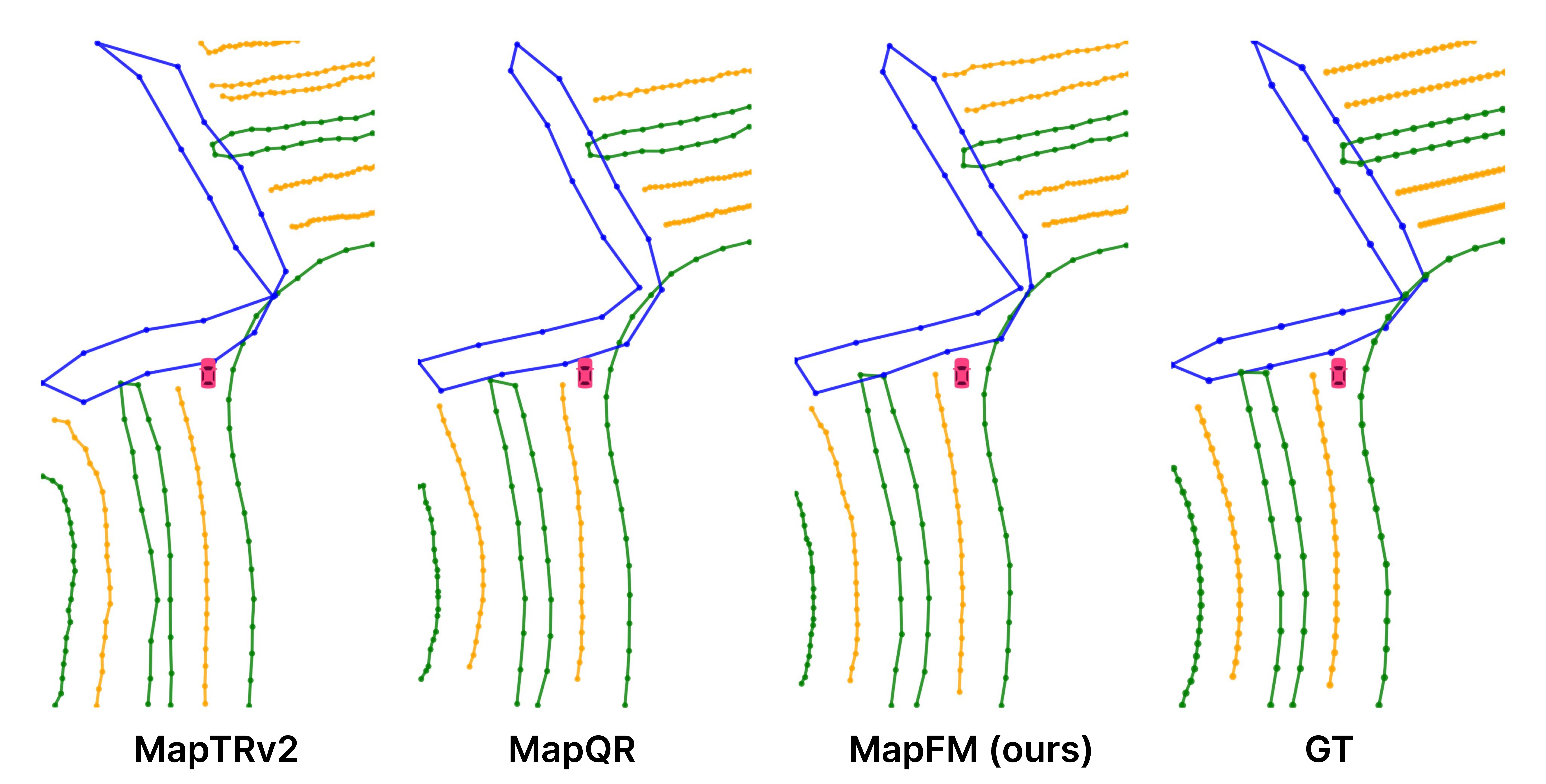}
  \end{subfigure}
  \caption{The qualitative comparison results of MapFM, MapQR~\cite{liu2024leveraging}, MapTRv2~\cite{liao2024maptrv2} on nuScenes val set.}
  \label{fig:qual_res}
\end{figure}

We show the overall evaluation results on NuScenes in Table \ref{tab:sota_compare}. 
Specifically, when using DINOv2-small as the backbone and training for 24 epochs, our MapFM achieves an mAP of 67.8\%. This represents a notable improvement of 1.5\% mAP over the MapQR (baseline) with a ResNet50 backbone 
(66.3\% mAP) and 1.7\% mAP over MapQR (baseline) with a SwinT backbone (66.1\% mAP). Furthermore, MapFM outperforms MGMapNet (ResNet50, 66.8\% mAP) by 1.0\% mAP under comparable training epochs. The advantages of MapFM become even more pronounced with a stronger backbone. Our MapFM equipped with DINOv2-base, despite being trained for only 16 epochs, reaches an impressive mAP of 69.0\%. This score significantly surpasses all other listed methods, including the MapQR baselines and MGMapNet, highlighting the effectiveness of integrating DINOv2 features with our proposed feature map aggregation and auxiliary road surface segmentation. Qualitative results are depicted in Fig. \ref{fig:qual_res}.


\subsection{Ablation Study}

\textbf{Auxiliary Road Surface Segmentation Head.} To evaluate the effectiveness of incorporating scene context understanding via multi-task learning, we investigate the impact of adding an Auxiliary Road Surface Segmentation Head to our baseline MapQR model~\cite{liu2024leveraging}. As discussed in Section 3.4, this auxiliary head is trained to predict a BEV mask representing the drivable road surface area, providing additional supervision.
\begin{table}[t]
\centering
\scriptsize
\caption{Comparison baseline with Auxiliary Road Surface Segmentation Head on nuScenes}
\label{tab:seg_head_comparison}
\begin{tabularx}{\linewidth}{|X|X|c|c|c|c|}
\hline
\textbf{Method} & \textbf{Backbone} & $\mathrm{AP}_{div}$ & AP$_{ped}$ & AP$_{bound}$ & \textbf{mAP} \\ \hline
MapQR (baseline)~\cite{liu2024leveraging} & ResNet50 & 68.0 & 63.4 & 67.3 & 66.3 \\ \hline
\textbf{MapQR + Aux Seg Head} & ResNet50 & \textbf{68.8} & \textbf{64.9} & \textbf{67.6} & \textbf{66.7} \\ \hline
MapQR (baseline)~\cite{liu2024leveraging} & SwinT & 68.1 & 63.1 & 67.1 & \underline{66.1} \\ \hline
\textbf{MapQR + Aux Seg Head} & SwinT & 68.1 & \textbf{64.1} & \textbf{68.0} & \textbf{67.1} \\ \hline
\end{tabularx}
\vspace{0.2cm}
\end{table}

The results in Table \ref{tab:seg_head_comparison} clearly demonstrate the benefit of the auxiliary road surface segmentation task. With the ResNet50 backbone, adding the segmentation head yields a consistent improvement, increasing the mAP from 66.3\% for the baseline MapQR to 66.7\%. This suggests that explicitly supervising the model to understand the road plane helps refine features for vectorized map element prediction.

More notably, the improvement is significantly more pronounced with the SwinT backbone. While the baseline MapQR with SwinT achieved 66.1\% mAP, adding the segmentation head boosts performance substantially to 67.1\% mAP, a 1.0 percentage point absolute improvement. This enhancement establishes the SwinT version of MapQR + Aux Seg Head as the top-performing configuration among those compared in this specific ablation, showcasing a strong synergy between the SwinT architecture and the auxiliary road segmentation task within our framework.

Overall, these comparisons validate our hypothesis: integrating an auxiliary task focused on understanding the broader spatial context, specifically the road surface area, serves as effective guidance for the BEV encoder. 

\textbf{Foundation Model Backbone.} The integration of powerful foundation models as backbones has shown significant promise in various vision tasks. In our MapFM approach, we leverage DINOv2~\cite{oquab2023dinov2} for its strong image representation capabilities. To understand its impact and the synergy with our proposed auxiliary supervision, we perform an ablation study, with results presented in Table~\ref{tab:seghead_for_mapfm}.
\begin{table}[t]
\centering
\scriptsize
\caption{Ablation Study on Auxiliary Road Surface Segmentation Head for MapFM with DINOv2}
\label{tab:seghead_for_mapfm}
\begin{tabular}{|l|c|c|c|c|c|c|}
\hline
\textbf{Configuration} & \textbf{Backbone} & \textbf{Epochs} & $\mathrm{AP}_{div}$ & AP$_{ped}$ & AP$_{bound}$ & \textbf{mAP} \\ \hline
MapFM w/o Aux Seg Head & DINOv2-small & 14 & 62.2 & 60.8 & 66.1 & 63.0 \\ \hline
MapFM w/ Aux Seg Head & DINOv2-small & 14 & \textbf{63.7} & \textbf{62.3} & \textbf{66.4} & \textbf{64.1} \\ \hline
\end{tabular}
\vspace{0.2cm}
\end{table}

The results clearly demonstrate that even with a strong foundation model like DINOv2 providing rich initial features, the introduction of the auxiliary road surface segmentation task provides valuable contextual guidance. This auxiliary supervision forces the model to learn more discriminative BEV features relevant to road layout and drivable space, which in turn benefits the primary vectorized map element prediction task.

To optimize the representations from the DINOv2 backbone, we evaluated feature aggregation from multiple blocks (4, 8, and 12) using various strategies, including feature concatenation and a multi-layer CNN with lateral connections to blocks (see Table \ref{tab:dino_feat_agr}). Our findings show that, for this task and architecture, using features solely from the last DINOv2 block yielded the best performance. Both other approaches have equally performed worse.

\begin{table}[t]
\centering
\scriptsize
\caption{Ablation Study on DINOv2 Feature Aggregation Strategies}
\label{tab:dino_feat_agr}
\begin{tabular}{|l|c|c|c|}
\hline
\textbf{Feature Aggregation Strategy} & \textbf{Backbone} & \textbf{Epochs} & \textbf{mAP} \\ \hline 
Multi-layer CNN (4, 8, 12) & DINOv2-small & 12 & 62.6 \\ \hline
Feature concatenation (4, 8, 12) & DINOv2-small & 12 & 63.9 \\ \hline
Last Layer Features (12) & DINOv2-small & 12 & \textbf{64.1} \\ \hline
\end{tabular}
\vspace{0.2cm}
\end{table}

Also we investigated the effectiveness of different transfer learning strategies using pre-trained foundational models, specifically RADIOv2.5 and DINOv2, as backbones for our HD map generation task. Our ablation study (see Table \ref{tab:dino_transfer_learning}) explored both frozen feature extraction and fine-tuning approaches.
\begin{table}[t]
\centering
\scriptsize
\caption{Ablation Study on Foundation Model Transfer Learning Strategies}
\label{tab:dino_transfer_learning}
\begin{tabular}{|l|c|c|c|c|c|}
\hline
\textbf{Transfer Learning Strategy} & \textbf{Epochs} & $\mathrm{AP}_{div}$ & AP$_{ped}$ & AP$_{bound}$ & \textbf{mAP} \\ \hline
Frozen RADIOv2.5 & 14 & 56.4 & 53.8 & 61.6 & 57.3 \\ \hline
Frozen DINOv2 & 14 & 49.2 & 50.2 & 57.7 & 52.4 \\ \hline
Fine-tune RADIOv2.5's Last Layer & 14 & 61.5 & 57.8 & 62.9 & 60.7 \\ \hline
Fine-tune DINOv2's Last Layer & 14 & \textbf{66.0} & \textbf{62.8} & \textbf{66.7} & \textbf{65.2} \\ \hline
\end{tabular}
\vspace{0.2cm}
\end{table}

Frozen RADIOv2.5 demonstrated a stronger baseline performance compared to DINOv2. This suggests that the features learned by RADIOv2.5 in its original training are, to some extent, more readily transferable or aligned with the requirements of our downstream task without further adaptation. However, a more significant performance gain was observed with DINOv2 when a fine-tuning strategy was applied, even when limited to just its last layer. This approach substantially outperformed both frozen model configurations.


\section{Conclusion}
In this paper, we presented significant improvements by incorporating powerful foundation models like DINOv2 for image encoding and by integrating auxiliary BEV semantic segmentation heads. This multi-task approach provides richer supervision, leading to more comprehensive scene representation and higher accuracy in predicted vectorized HD maps.

\textbf{Limitations.}
While foundation models offer significant benefits, their direct, out-of-the-box application for this task remains challenging without fine-tuning. Achieving strong performance still necessitates a degree of task-specific adaptation.

\textbf{Future Work.}
We plan to explore other foundation models and investigate more advanced multi-scale feature aggregation strategies from these models. 

\section*{Acknowledgements}
This work was supported by the
Ministry of Economic Development of the Russian Federation (code 25-139-66879-1-0003).


%
%
%
\bibliographystyle{splncs04}
\bibliography{bibliography}

\end{document}